\documentclass[acmtog]{acmart}

\copyrightyear{2025}
\acmYear{2025}
\setcopyright{cc}
\setcctype{by}
\acmConference[SIGGRAPH Conference Papers '25]{Special Interest Group on Computer Graphics and Interactive Techniques Conference Conference Papers }{August 10--14, 2025}{Vancouver, BC, Canada}
\acmBooktitle{Special Interest Group on Computer Graphics and Interactive Techniques Conference Conference Papers (SIGGRAPH Conference Papers '25), August 10--14, 2025, Vancouver, BC, Canada}
\acmDOI{10.1145/3721238.3730609}
\acmISBN{979-8-4007-1540-2/2025/08}

\acmSubmissionID{251}
\usepackage{url}
\usepackage{subfig}
\usepackage{wrapfig}

\usepackage{glossaries}
\newacronym{csf}{CSF}{contrast sensitivity function}
\newacronym{cpd}{cpd}{~cycles per degree}
\newacronym{ipd}{IPD}{interpupillary distance}
\newacronym{vr}{VR}{virtual reality}
\newacronym{ar}{AR}{augmented reality}
\newacronym{hvs}{HVS}{human visual system}
\glsdisablehyper

\graphicspath{{fig/}}

\begin{document}

\title{Towards Understanding Depth Perception in Foveated Rendering} 

\author{Sophie Kerga{\ss}ner}
\email{sophie.kergassner@usi.ch}
\affiliation{\institution{Università della Svizzera italiana}\country{Switzerland}}

\author{Taimoor Tariq}
\email{taimoor.tariq@usi.ch}
\affiliation{\institution{Università della Svizzera italiana}\country{Switzerland}}

\author{Piotr Didyk}
\email{piotr.didyk@usi.ch}
\affiliation{\institution{Università della Svizzera italiana}\country{Switzerland}}

\begin{abstract}

The true vision for real-time virtual and augmented reality is reproducing our visual reality in its entirety on immersive displays. To this end, foveated rendering leverages the limitations of spatial acuity in human peripheral vision to allocate computational resources to the fovea while reducing quality in the periphery. Such methods are often derived from studies on the spatial resolution of the human visual system and its ability to perceive blur in the periphery, enabling the potential for high spatial quality in real-time. However, the effects of blur on other visual cues that depend on luminance contrast, such as depth, remain largely unexplored. It is critical to understand this interplay, as accurate depth representation is a fundamental aspect of visual realism. 
In this paper, we present the first evaluation exploring the effects of foveated rendering on stereoscopic depth perception. We design a psychovisual experiment to quantitatively study the effects of peripheral blur on depth perception. Our analysis demonstrates that stereoscopic acuity remains unaffected (or even improves) by high levels of peripheral blur. Based on our studies, we derive a simple perceptual model that determines the amount of foveation that does not affect stereoacuity. Furthermore, we analyze the model in the context of common foveation practices reported in literature. The findings indicate that foveated rendering does not impact stereoscopic depth perception, and stereoacuity remains unaffected with up to $2\times$ stronger foveation than commonly used. Finally, we conduct a validation experiment and show that our findings hold for complex natural stimuli.

\end{abstract}

\ccsdesc[500]{Computing methodologies~Perception}
\ccsdesc[500]{Computing methodologies~Virtual reality}

\keywords{Human Perception, Stereoacuity, Perceptual Model, Foveated Rendering}

\begin{teaserfigure}
\includegraphics[width = \linewidth]{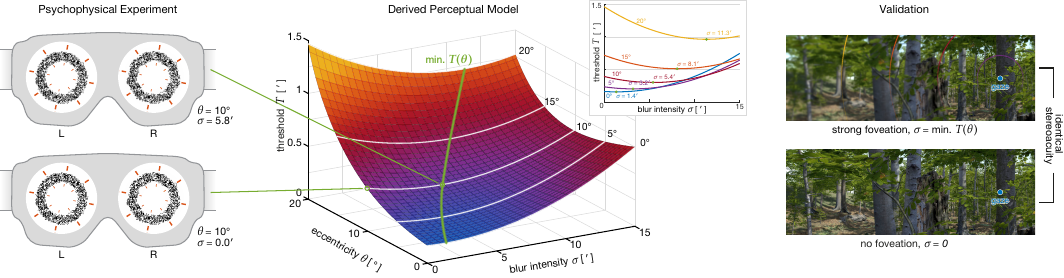}
\caption{
We design a broadband stimulus to accurately estimate disparity thresholds across various eccentricities and blur levels.
We fit a custom surface through the retrieved threshold estimations. It shows the lowest resolvable disparity $T$ as a continuous function of eccentricity $\theta$ and Gaussian filter sigma $\sigma$.
The green line denotes the minimum $T$-value per eccentricity $\theta$, encoding the amount of blur that can be induced by foveated rendering without degrading stereoscopic depth perception.
We validate our findings in a validation study, by showing participants unaltered renderings vs. strongly foveated renderings.
}
\label{fig:mod-surface}
\Description{ }
\end{teaserfigure}

\maketitle

\section{Introduction}

The \gls{hvs} exhibits a non-uniform sensitivity towards different visual cues across the visual field. While visual acuity, shape discrimination, and sensitivity to color and depth corrugations are highest in the central fovea, there is a substantial decline in sensitivity to these factors with increasing eccentricity \cite{rosenholtz-2016a,whitaker-1993a,baldwin-2016a,abramov-1991a,moreland-1959a,prince-1998}. This is the prime motivation behind foveated rendering techniques, that allocate computational resources to the fovea while reducing fidelity in the periphery. With the advent of gaze-tracking systems in \gls{ar} and \gls{vr}, foveated rendering has emerged as a key enabler of real-time high-quality rendering for virtual realities. Consequently, it is being incorporated by graphics APIs, such as Unity and Nvidia VRS, and used by commercial \gls{ar} and \gls{vr} devices, such as Meta Quest and Apple Vision Pro.

The basis for foveated rendering methods are studies and psychophysical models of peripheral vision \cite{strasburger2011}, which form the foundation for perceptually optimized rendering methods \cite{murphy2001,guenter-2012a,patney-2016a, tursun-2019a, surace-2023a}. However, while they focus on different aspects of spatial quality, they neglect the potential influence of foveated rendering on visual cues such as motion \cite{tariq-2024a} and depth \cite{sun-2019a}. Consequently, the effects of foveated rendering on more general human perception, which governs the interaction and navigation in three-dimensional environments, remain largely unexplored.
This is particularly relevant given the availability and importance of depth cues in novel displays, which inherently utilize foveated rendering.

In this work, we address the problem of reproducing and preserving depth cues on novel displays. While innovative display designs have already highlighted their importance \cite{akeley-2004a,akcsit-2017a, Narain-2015a,qin-2023a}, our work addresses the problem from the rendering perspective. We take first steps towards understanding how foveated rendering influences stereopsis, one of the most fundamental depth cues \cite{cutting-1995a}. While past research has already demonstrated a decrease in stereoacuity due to eccentricity \cite{prince-1998}, as well as the significant impact of blur on stereopsis \cite{cormack1991}, these effects have not yet been investigated in conjunction, as occurring in foveated rendering. To close this gap, we design a broadband stimulus that serves as a basis for our psychophysical experiment (Section \ref{sec:methods}) investigating the effect of blur intensity on stereopsis at different eccentricities. Based on the measurements, we derive a model (Section \ref{sec:model}) that describes the maximum blur intensity which does not compromise stereopsis. We discuss the findings in the context of common practices in foveated rendering literature. Our findings suggest that stereopsis in peripheral vision is remarkably resilient to loss of spatial resolution, offering new possibilities for resource-efficient rendering strategies without sacrificing spatial awareness. Finally, we validate our findings by applying our model to realistic scenes (Section \ref{sec:validation}).
\section{Background and Previous Work}\label{sec:background}

Decoding human vision has received the efforts of many. In this work, we focus in particular on the perception of stereoscopic depth, and the effects of peripheral blur on stereoacuity.

\subsection{Peripheral Vision and Foveated Rendering}

Peripheral vision exhibits significantly reduced spatial resolution compared to foveal vision \cite{rosenholtz-2016a}. This is primarily due to the lower density of cone photoreceptors and ganglion cells in the retina's periphery \cite{curcio-1990a}. While the fovea supports high-acuity tasks, the periphery excels at motion detection \cite{mckee-1984a} and spatial learning \cite{fortenbaugh-2008-EffectPeripheralVisual,alfano-1990-RestrictingFieldView}. Foveated rendering utilizes these perceptual limitations by reducing rendering quality in peripheral regions. While researchers initially focused on reducing spatial resolution in the periphery \cite{guenter-2012a}, subsequent research investigated pre- and post-processing steps which allowed for more aggressive foveation. \citeauthor{patney-2016a} [\citeyear{patney-2016a}] post-applied contrast enhancement, allowing higher blur rates. Similarly, \citeauthor{tariq-2022a} [\citeyear{tariq-2022a}] recovered detectable frequencies in a cost-efficient post-processing step to re-enhance foveated content. \citeauthor{tursun-2019a} [\citeyear{tursun-2019a}] modeled the masking effects of the underlying content, enabling content-aware foveation. \citeauthor{surace-2023a} [\citeyear{surace-2023a}] examined substantially reduced geometric complexity in the periphery. Another approach featured inducing mental tasks in the fovea to allow for an unnoticed increase in foveation. \cite{krajancich-2023a}. However, recent work on motion perception revealed a damage of speed estimations induced by foveation and resolved this by recovering spatio-temporal energy in a post-processing step \cite{tariq-2024a}.

\subsection{Depth Perception}

To navigate 3D space, humans rely on a number of different depth cues that convey either  relative or absolute distance information \cite{sweet-2012a,cutting-1995a}. Cues such as accommodation, convergence, or familiar size convey information about the absolute distance between the observer and an object. Other cues provide solely relative information of ordinal or metric kind, allowing for the estimation of relative depth differences between objects. For example, occlusion conveys only ordinal information, whereas cues such as stereoscopic disparity or relative size provide metric information regarding the magnitude of depth differences.

With the advent of stereoscopic displays, the recreation of binocular vision became feasible. This enabled the display of stereo-
\setlength{\columnsep}{14pt}
\begin{wrapfigure}[9]{r}{0.45\linewidth}
    \centering
    \includegraphics[width=\linewidth]{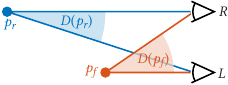} 
    \caption{Different vergence angles occurring for different points.}
    \label{fig:bpw-disparity}
    \Description{}
\end{wrapfigure}
scopic depth cues such as binocular disparity, resulting in stereopsis. Stereopsis is created by perceiving a difference in vergence angles between a focus point $p_f$ and a reference point $p_r$, which naturally occurs when those points are at different depths with respect to the eyes (Figure \ref{fig:bpw-disparity}). The vergence angles can be described by $D(p) = \angle L p R$, denoting the angles between the two locations of the eyes in space ($L$ and $R$ ), and the respective points $p_f$ and $p_r$. Stereopsis is quantified by the disparity $d$ [$^\circ$], denoting the difference of these vergence angles: 
\begin{equation}\label{eq:disparity}
    d = D(p_1) - D(p_2).
\end{equation}
The sign of $d$ denotes whether objects are relatively closer, or farther. In case both vergence angles are the same, $d=0$ and no disparity is present. This can happen when both points are at the same position in space, the image is rendered with an \gls{ipd} of 0mm, or both points lie on the horopter \cite{blakemore-1970a}.

Large population studies found average disparity thresholds of 30--60$''$ disparity, with lowest thresholds as low as 2$''$ \cite{coutant-1993,bosten-2015-PopulationStudyBinocular}. However, stereoacuity depends strongly on various properties exhibited by the luminance pattern. \citeauthor{siderov-1995a} [\citeyear{siderov-1995a}] studied band-limited luminance patterns, finding peak sensitivities at 8\gls{cpd} in the fovea, and 2\gls{cpd} at $10^\circ$ eccentricity. \citeauthor{didyk-2012a} [\citeyear{didyk-2012a}] manipulated disparity based on the underlying luminance contrast. Their perceptual model is based on a finding of \citeauthor{marr-1979a} [\citeyear{marr-1979a}], who claim that disparity is mostly supported by the most sensitive channel. Additionally, overall stereopsis declines significantly for increasing eccentricities \cite{mochizuki-2012a,prince-1998}. In addition to the factors mentioned above, stereoacuity is affected by the characteristics of the displayed depth modulation. \citeauthor{bradshaw-1999a} [\citeyear{bradshaw-1999a}] found that humans are more sensitive to horizontally than to vertically oriented corrugations, and are particularly sensitivity to depth modulations of 0.3--0.5\gls{cpd}. This finding was reproduced by \citeauthor{didyk-2011a} [\citeyear{didyk-2011a}]. \citeauthor{prince-1998} [\citeyear{prince-1998}] measured the peak sensitivity to spatial corrugation frequencies in the periphery and found that the sensitivity M-scales.

The endeavor of accurately reproducing cues of depth and focus in \gls{ar} and \gls{vr} has also been a subject of strong efforts from a display design perspective. An early multi-focal display was presented by \citeauthor{akeley-2004a} [\citeyear{akeley-2004a}], which delivered the important depth cue of accommodation. \citeauthor{yang-2023a} [\citeyear{yang-2023a}] introduced the concept of varifocal \gls{vr}, which has a movable display to accurately deliver vergence cues to users in addition to disparity. Recently, Split-Lohmann multifocal displays demonstrated the ability to deliver a dense set of focal planes simultaneously using a single exposure \cite{qin-2023a}.

While all of the aforementioned studies and efforts address various aspects of human depth perception at the fovea, only few of them explore the peripheral vision, which forms \textasciitilde95\% of our visual field. Although we rely on foveal vision for many tasks, reconstructing a realistic depth impression across the entire visual field is essential, particularly as the periphery plays a critical role in accurately understanding spatial layout \cite{fortenbaugh-2008-EffectPeripheralVisual,alfano-1990-RestrictingFieldView}. To the best of our knowledge, human disparity perception under the influence of peripheral blur is an open question. Given the advent of gaze-tracked \gls{vr}, it is critical to understand how techniques such as foveated rendering affect depth perception.
\section{Methods}\label{sec:methods}

We aim to design a perceptual model that quantifies the stereoscopic acuity threshold $T(\theta,\sigma)$ (minimum disparity that is resolvable) as a function of blur intensity $\sigma$ and eccentricity $\theta$. For this, we design an experiment displaying a textured ring with a sinusoidal depth corrugation of varying amplitude. The texture is broadband, to quantitatively study the effect of the loss of spatial frequencies on depth perception. We adjust the depth amplitude, and thus the disparity $d$ between peaks and troughs, in a 2-Alternative Forced Choice staircase procedure to determine the resolvability threshold. We measure the stimulus at three eccentricities: 0$^\circ$ (fovea), 10$^\circ$, and 20$^\circ$ eccentricity (periphery). The upper measurable eccentricity is limited by headset components, which allow binocular vision only up to $25^\circ$ eccentricity. At each eccentricity, we blur the stimulus with a Gaussian filter of varying size. The filter's $\sigma$ varies from $0'$--$15'$. For 20$^\circ$ eccentricity, we include an additional measurement for $\sigma = 26.6'$ (see Table \ref{tab:met-stimuli-grid} for an overview).

\begin{table}
    \caption{Tested permutations of the independent variables $\theta$ and $\sigma$. $\sigma$ is constant in pixel-scale, thus corrected for $\theta>0$.}
    \begin{tabular}{l *7{p{19.2pt}}}
        \toprule
        &\multicolumn{7}{c}{$\boldsymbol{\rightarrow \quad}$ \textbf{increasing} $\boldsymbol{\sigma~['] \quad \rightarrow}$}\\
        \midrule
        $\boldsymbol{\theta=0^\circ}$&
        $0.0'$& 
        $3.0'$& 
        $6.0'$&
        $9.0'$&
        $12.0'$&
        $15.0'$&
        $\times$\\
        $\boldsymbol{\theta=10^\circ}$&
        $0.0'$&
        $2.9'$&
        $5.8'$&
        $8.7'$&
        $11.6'$&
        $14.6'$&
        $\times$\\
        $\boldsymbol{\theta=20^\circ}$&
        $0.0'$&
        $2.6'$&
        $5.3'$&
        $8.0'$&
        $10.6'$&
        $13.3'$&
        $26.6'$\\
        \bottomrule
    \end{tabular}
    \label{tab:met-stimuli-grid}
\end{table}

The stimulus design minimizes the influence of all depth cues except binocular disparity.
While natural content presents a variety of depth cues such as motion parallax, shading, texture density, and deformation, the purpose of this study is to particularly measure disparity sensitivity in isolation. To this end, we ensure that monocular observation of the stimulus does not convey depth information, except for minor texture deformation. However, the visibility of the texture deformation is well below the resolvable disparity limit and, therefore, negligible \cite{bradshaw-1999a}.

\subsection{Stimulus Design}

\begin{figure}
    \includegraphics[width=\linewidth]{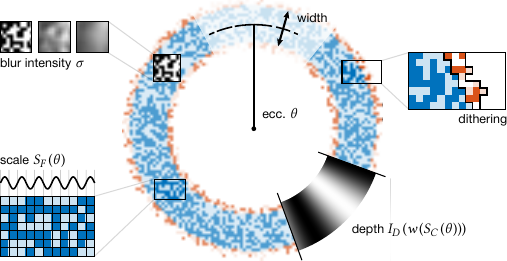}
    \caption{Overview of the controlled stimulus parameters: eccentricity $\theta$, the ring's width, blur intensity $\sigma$, texture scale based on $S_F(\theta)$, depth map $I_D$ based on $w(S_C(\theta))$, and the dithered border.}
    \label{fig:met-stimulus-parameters}
    \Description{}
\end{figure}

We design our stimulus in the shape of a ring.
Using a ring has the advantage of measuring the whole visual field uniformly, independent of the distinctive characteristics and strengths exhibited by different visual directions. The ring exhibits a variety of properties, including luminance pattern, and modulated depth (Figure \ref{fig:met-stimulus-parameters}).
We control every parameter in a way that optimizes it for human vision. This is an effort to try to measure the lowest possible stereoacuity threshold. Given the numerous additional factors that influence depth perception that we do not control, such as presentation time, participant training, eye movement, or overall luminance level \cite{wolski-2022a,fendick-1983a,rady-1955a}, it is important to acknowledge that our measured thresholds may not accurately reflect the absolute threshold. Instead, they provide a general trend and demonstrate the behavior of stereoacuity under different conditions. The respective design choices are motivated in the following sections.

\subsubsection{Texture}

50\% Random Dot Patterns are widely used in vision science, particularly for researching depth perception \cite{julesz-1960a}. They serve as a simple baseline pattern for our metric, exhibiting controllable features such as scale, frequency range, and dynamic range. As we aim to measure the lowest possible disparity threshold, we scale the random dot pattern to the optimally perceptible size. This scale equals the peak sensitivity value of the \gls{csf} for the given eccentricity $\theta$. The stelaCSF\footnote{\url{https://github.com/gfxdisp/stelaCSF}, retrieved 18. Oct 2024} \cite{stelaCSF} yields a peak sensitivity $S_F(\theta)$ of $S_F(0^\circ)=4.1$\gls{cpd}, $S_F(10^\circ)=1.8$\gls{cpd}, and $S_F(20^\circ)=1.3$\gls{cpd}. We scale the dot pattern to match $S_F(\theta)$, by scaling the dots such that two dots equal one cycle.

Our setup allows us to accurately pre-blur the ring's texture to precisely measure each $\sigma$ level. We pre-render all required blur intensities by using $\sigma$ as the standard deviation of a Gaussian filter kernel. After blurring, we recover the full dynamic range, as a reduced contrast might make it more challenging to stereo match the images and thus perceive the depth. This would result in a loss of comparability between stimuli with varying $\sigma$.

Additionally, the ring must be of a sufficient minimum width in the visual field to ensure good visibility. However, to reduce crosstalk between eccentricities and ensure accurate measurements, the width is minimized. To account for the varying sensitivity per eccentricity, we scale the ring wider with higher eccentricities. We set the ring's width based on the \gls{csf}, choosing a width of 13 dots for both peripheral stimuli. This is approximately equal to $3.8^\circ$ in the visual field for $10^\circ$, and $5^\circ$ for $20^\circ$ eccentricity. For $0^\circ$ eccentricity we set a width of $6.7^\circ$, so that always 2 full cycles are visible. Additionally, the ring has a randomly dithered border (Figure \ref{fig:met-stimulus-parameters}). This attenuates the monoscopic depth cue of geometric deformation, which occurs due to the pixel shifting.

\subsubsection{Depth}

Similarly to the spatial frequency of the texture, we display the ring with a depth corrugation frequency that humans are most sensitive to. In the fovea, humans have the highest sensitivity to depth corrugations with a spatial frequency of 0.2--0.5\gls{cpd} \cite{bradshaw-1999a,didyk-2011a}. However, the ideal spatial frequency for depth corrugations shifts with eccentricity. According to \citeauthor{prince-1998} [\citeyear{prince-1998}], the sensitivity M-scales with 0.08 cycles per mm of cortex. This value is an average of the sensitivities across all directions of the visual field, and is therefore evenly applicable to our ring. We extract the M-values as an average across the four directions provided from \cite{rovamo-1979}, and retrieve the peak sensitivity $S_c(\theta)$, with $S_c(10^\circ) = 0.133$\gls{cpd}, and $S_c(20^\circ) = 0.073$\gls{cpd}. For $S_c(0^\circ)$ we set a value of $0.3$\gls{cpd}. Based on the circumference of the ring, we determine the number of depth corrugation cycles $w(\theta)$ per eccentricity. This yields $w(10^\circ) = 8$, and $w(20^\circ) = 9$. The depth maps $I_D(w(\theta))$ are pre-rendered sinusoidal patterns (as shown in Figure \ref{fig:met-stimulus-parameters}). For $0^\circ$ they are horizontally oriented \cite{bradshaw-1999a}, for the other cases they are radially oriented. For each eccentricity, we pre-render five depth maps with a phase shift of $+0.4\pi$ each for later randomization.

To induce stereopsis, we artificially shift a pixel $p_R$ in the right 
\setlength{\columnsep}{14pt}
\begin{wrapfigure}[11]{r}{0.55\linewidth}
    \includegraphics[width=\linewidth]{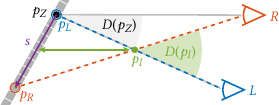} 
    \caption{Schematic representation of the induced disparity between reference point $p_Z$ and artificially created point $p_I$.}
    \label{fig:creating-disparity}
    \Description{}
\end{wrapfigure}
eye's image, while leaving it untouched in the left eye's image ($p_L$). This procedure follows implementations of previous work \cite{siderov-1995a,bradshaw-1999a,didyk-2011a} and guarantees precise control of the presented disparity. The pixel shift introduces an intersection point $p_I$ which our brain perceives as being in front of the zero-plane (Figure \ref{fig:creating-disparity}). The disparity is measured between the point $p_Z$, which serves as a zero point reference (trough, $I_D(p_Z)=0$), and $p_I$, the artificially created illusion point (peak, $I_D(p_I)=1$). We generate the right eye's image $I_R$ by resampling the left eye's image $I_L$ at the respective image coordinates:
\begin{equation}\label{eq:resampling}
    I_R(u,v) = I_L(u + s(d)\cdot I_D(u,v), v),
\end{equation}
with $s(d)$ being the shift amount in UV-coordinates based on the desired disparity $d$. The shift $s$ is weighed by the depth map $I_D$ to create the corrugations. This procedure results in a wavily warped image $I_R$, as each pixel is sampled slightly shifted from $I_L$ if $I_D>0$. The final stimuli are shown in Figure \ref{fig:met-final-stimuli}. An exemplary stereogram is provided in the Supplementary Material.

\begin{figure}
    \includegraphics[width=.31\linewidth, page = 1]{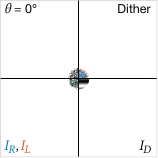}\hfill
    \includegraphics[width=.31\linewidth, page = 2]{fig/stimuli-final.pdf}\hfill
    \includegraphics[width=.31\linewidth, page = 3]{fig/stimuli-final.pdf}
    \caption{Our three final stimuli for $\theta \in \{0^\circ, 10^\circ, 20^\circ\}$.  The dithered border, as well as a representative depth map $I_D$ are shown in the right halves. The lower left shows the distortion in $I_R$ (blue) based on $I_D$.}
    \label{fig:met-final-stimuli}
    \Description{}
\end{figure}

\subsection{Psychophysical Experiment} \label{sec:met-user-study}

This section details the experimental protocol. One complete session covered the judgment of all 19 conditions listed in Table \ref{tab:met-stimuli-grid}.

\subsubsection{Procedure} \label{sec:met-procedure}

\begin{wrapfigure}[12]{r}{0.3\linewidth}
    \centering
    \includegraphics[width=\linewidth]{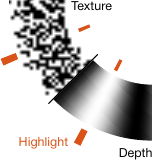} 
    \caption{The highlights mark the peaks.}
    \label{fig:mod-highlight}
    \Description{}
\end{wrapfigure}

For each condition, the threshold estimation was conducted as follows: Subjects viewed the stimulus with a highlight either marking the peaks, or the troughs (Figure \ref{fig:mod-highlight}). Each stimulus presentation was triggered by the participant and was visible for 1.5 seconds. This duration was chosen to have adequate stimuli display time for the $20^\circ$ eccentricity stimulus, as shorter presentation times 
\setlength{\columnsep}{14pt}
led to higher variability in measurements across different trials, as shown in pilot experiments. Pilot testing also showed that unlimited presentation times lead to visual artifacts, such as strong afterimages, during observation. Participants were asked whether the highlight marks the peaks or the troughs. After the presentation, they logged their answers via keyboard input. The stimulus intensity (amount of disparity $d$) was determined by the "Best PEST" procedure \cite{pentland-1980,lieberman-1982}. After a maximum of 60 stimulus presentations per condition, the PEST procedure was stopped and the data was extracted.

\begin{figure*}
    \includegraphics[width=\linewidth]{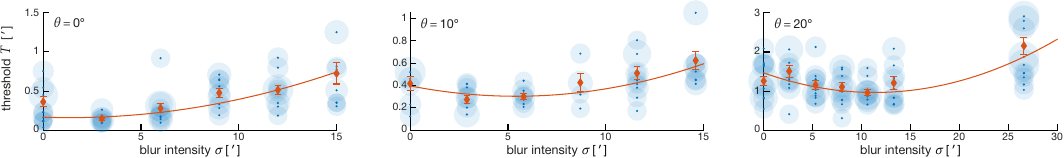}
    \caption{Plot of estimated thresholds per eccentricity, as well as the polynomial fits. Each blue point denotes one measured threshold, the halo of each point denotes its weight $w$. The orange scatter plots denote the mean and standard error per $\sigma$. The line plots show the retrieved parabolas, fitted through the weighted threshold data.}
    \label{fig:mod-parabolas}
    \Description{}
\end{figure*}

\subsubsection{Apparatus} \label{sec:met-apparatus}

The study was conducted on a Varjo XR-3 headset. We deactivated the focus displays in the Varjo Base Application and only presented the stimulus on the context displays (2880$\times$2720px per eye). The application was created in Unity Version 2022.3.9 using OpenGL shader. Since all resource-intensive calculations are precalculated, we operated in real-time at 90 frames per second constantly.

\subsubsection{Participants} 

The threshold estimation of $\theta=10^\circ$ and $\theta=20^\circ$ was conducted on 11 voluntary participants (1F, 10M), with an average age of 28 ± 4.7 years. All participants had normal or corrected-to-normal vision and were tested negative for stereo blindness. Nine participants reported prior experience in conducting perceptual experiments. The threshold estimation of $\theta=0^\circ$ was conducted on 10 of the 11 previous participants (1F, 9M; mean age 28.1 ± 4.9 years). Eight participants reported prior experience. With the exception of the two participating authors, all participants were na\"{\i}ve to the project. Additionally, all participants signed their previous consent and were compensated for their time. The study was approved by the institutional ethics committee.
\section{Results and Perceptual Model}\label{sec:model}

This section covers the evaluation of the psychophysical experiment to retrieve all thresholds $T_p(\theta,\sigma)$ per participant and condition, as well as the design and fitting of our perceptual model.

\subsection{Threshold Estimation}\label{sec:mod-threshold-estimation}

The PEST procedure yields one dataset per measured condition which is evaluated independently to extract the estimated threshold $T_p(\theta,\sigma)$. We generate a data point for each measured disparity intensity $d$, such that the value denotes the probability of detection as $P(d) = \frac{c_{d}}{n_{d}}$, where $n_{d}$ is the number of samples per disparity, and $c_{d}$ is the number of correctly detected stimuli. While a detection probability of 50\% means that the participant failed to resolve the stimulus and answered randomly, a 100\% score means that the participant correctly resolved the depth in each trial. We fit a psychometric curve and determine the 75\% point. The 75\% point is commonly used to describe an above-random score, which means that participants resolved the stimulus correctly. We fit a Weibull cumulative distribution function $F$, and estimate the threshold $T$
\begin{equation}
    x = T_p(\theta,\sigma) \text{~, such that~} F(x) = 0.75.
\label{eq:threshold}
\end{equation}
Since our probability values range from $P=0.5$ (random guess) to $P=1$ (always correct answer), we fit a modified Weibull curve that outputs values $F \in [0.5, 1]$:
\begin{equation}
    F(x) = 1 - \frac{1}{2} \cdot e^{-(\frac{x}{\lambda})^k}\text{,}
\label{eq:weibull2}
\end{equation}
where $\lambda$ and $k$ are the parameters we fit. For the fit, we weigh each data point $P(d)$ with the weight $w(d)=n_d$. This ensures that frequently measured disparity levels become a larger weight in the fit, as it is more certain that $P(d)$ describes the real value.

Since the PEST procedure can lead to sparse sampling of important regions and random answers can render data unreliable, the Weibull fit can get unstable. We obtain the final threshold $T$ as well as its standard deviation $T_{\sigma}$ by bootstrapping the data 100 times. This also yields a relative uncertainty of $u = \frac{T_{\sigma}}{T}$ per $T$. Measurements with $u > 0.3$ are treated as outliers and excluded from further evaluation, as we consider the PEST procedure to be too unstable and therefore unreliable. This affects \textasciitilde20\% of our measurements. A plot of the excluded data points is provided in the Supplementary Material.

\subsection{Perceptual Model}

\begin{figure}
    \includegraphics[width=\linewidth]{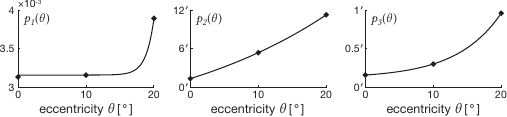}
    \caption{Fit of the three parabola parameters $p_1$, $p_2$, and $p_3$. The fitted curves are exponential curves of shape $p_i(\theta) = ae^{b\theta}+c$.}
    \label{fig:mod-parameter-pi}
    \Description{}
\end{figure}

We aim to fit a continuous surface $M$ through our acquired thresholds $T_p(\theta,\sigma)$. For this, we take a look at the characteristics exhibited by our measured thresholds (Figure \ref{fig:mod-parabolas}): Instead of increasing monotonically, the mean values of $T$ first decrease with increasing $\sigma$, before eventually rising again. To match this characteristic, we fit a parabola of the form $p_1(\theta)\cdot(\sigma-p_2(\theta))^2+p_3(\theta)$ to  each eccentricity $\theta$ (Figure \ref{fig:mod-parabolas}), where $p_1$ defines the steepness, $p_2$ the shift of the minimum in $\sigma$-direction, and $p_3$ denotes the $T$-value of the minimum of the respective parabola. Each data point is weighted by an inverse of its uncertainty $u$ with $w(u) = \frac{1}{u^2}$ to ensure that more reliable threshold estimations take more influence on the fit. From this, we interpolate between the three parabolas to complete the model and retrieve the surface $M$ of shape
\begin{equation}\label{eq:mod-surface-shape}
    M =  p_1(\theta)\cdot(\sigma-p_2(\theta))^2+p_3(\theta), \quad   \theta\in[0,20].
\end{equation}
To estimate the parameters' interpolation characteristics for varying $\theta$, we examine the characteristics of our fitted parabolas (Figure \ref{fig:mod-parabolas}): The minimum threshold, associated with $p_2$, shifts to a higher $\sigma$ with increasing eccentricity. This indicates that the \gls{hvs} tolerates higher blur levels in the periphery. Furthermore, the overall threshold $T$ shifts to higher values with increasing eccentricity, which resembles the y-offset encoded in $p_3$. The slope $p_1$ remains mostly unaltered. Figure \ref{fig:mod-parameter-pi} displays the parameter-values $p_i$ per eccentricity. As expected, they resemble the intuitive understanding discussed above. To model the parabolas for a continuous $\theta$, we fit an exponential curve of the form $p_i(\theta) = ae^{b\theta}+c$. An exponential curves increase strictly, which resembles our data. The fitting yields:

\begin{equation} \label{eq:mod-parameters}
    \begin{split}
        p_1(\theta) =~ & 2.07 \cdot 10^{-11} \cdot e^{0.87\cdot\theta}+0.003\\
        p_2(\theta) =~ & 8.85 \cdot e^{0.04\cdot\theta}-7.5\\
        p_3(\theta) =~ & 0.04 \cdot e^{0.15\cdot\theta}+0.12
    \end{split}
\end{equation}
The final surface fit $M$ is displayed in Figure \ref{fig:mod-surface}. It is described by inserting the fitted parameters $p_i$ in Equation \ref{eq:mod-surface-shape}, which results in:
\begin{equation}\label{eq:mod-model}
    \begin{split}
         M = &   \left[2.07 \cdot 10^{-11} \cdot e^{0.87\cdot\theta}+0.003\right] \cdot\\
            & \left(\sigma - \left[8.85 \cdot e^{0.04\cdot\theta}-7.5\right]\right)^2 +\\
             &\left[ 0.04 \cdot e^{0.15\cdot\theta}+0.12 \right]
    \end{split}
\end{equation}
This model holds for our measured scope of $\sigma\in[0, 15]$, and $\theta\in[0, 20]$. However, we note that $p_1$ exhibits a steep slope for higher eccentricities, resulting in extreme values for $\theta > 20$. If such conditions are expected we suggest replacing $p_1$ with a constant fit of $p_1 = 0.0034$, which is the mean across $p_1(0^\circ)$, $p_1(10^\circ)$ and $p_1(20^\circ)$. Replacing $p_1$ with this value only results in a minor difference to the current model, as the surface is still well-defined by $p_2$ and $p_3$, and $p_1$’s variability across the three measured eccentricities is not extreme.

\subsection{Discussion}
\begin{figure}
    \includegraphics[width=\linewidth]{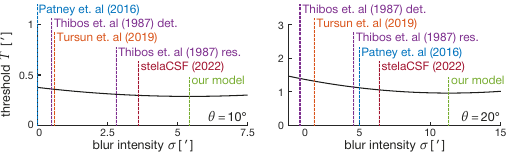}
    \caption{Our blur rates set into comparison with various $\sigma$-values reported in literature \cite{patney-2016a,tursun-2019a,thibos-1987a,stelaCSF}. If a cut-off frequency $f$[cpd] is reported, we retrieve $\sigma$ by assuming that the Gaussian filter has a kernel of size $\sigma = \frac{3}{2 \pi f}$.}
    \label{fig:mod-sigma-comparison}
    \Description{}
\end{figure}

We retrieve $p_2(\theta)$ as the blur level at which participants presented optimal stereoacuity. This yields exemplary $\sigma$-values of $p_2(10^\circ)= 5.41'$, and $p_2(20^\circ)= 11.33'$. Figure \ref{fig:mod-sigma-comparison} sets these values in comparison to foveation values reported in literature.
Our model exceeds all reported blur intensities, showing that stereoacuity is unaffected by commonly used levels of foveation. However, the shape of our model, especially the high tolerance for blur beyond the limit of visibility, is surprising. This means that stereopsis does not depend on perceivable high frequencies.

\begin{figure}
    \includegraphics[width=\linewidth]{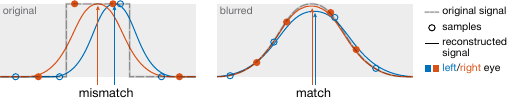}
    \caption{Illustration of how aliased frequencies influence stereo matching.}
    \label{fig:mod-theory}
    \Description{}
\end{figure}

Moreover, there seems to be a constant trend that the removal of high frequencies ($\sigma<p_2$) improves the stereoacuity slightly. We hypothesize that this is due to the removal of aliased frequencies: When sparsely sampling a continuous signal, only frequencies up to 0.5$\times$ sampling rate can be correctly reconstructed (Nyquist–Shannon sampling theorem). All frequencies above this threshold lead to aliasing. This means that the \gls{hvs} may interpolate the aliased samples incorrectly, leading to ambiguities (Figure \ref{fig:mod-theory}). This false recreation leads the \gls{hvs} to stereo match those signals incorrectly. Since correct stereo matching is the basis for correctly resolving disparity, this phenomenon leads to a distorted depth perception. Removing all confounding frequencies beforehand (low-pass filtering) prevents this mismatch. Figure \ref{fig:mod-sigma-comparison} marks the two equivalent $\sigma$-values that remove all aliased frequencies according to \citeauthor{thibos-1987a} [\citeyear{thibos-1987a}]. However, these $\sigma$-values are \textasciitilde50\% lower than our threshold. This contradicts the assumption that the trend of $T$ is solely influenced by the presence of aliased frequencies, since $T$ seems to be persistent beyond the removal of just the aliased frequencies. Instead, we hypothesize that there must be high frequencies that disturb stereo matching beyond the limit given by \citeauthor{thibos-1987a} [\citeyear{thibos-1987a}].

\section{Validation}\label{sec:validation}

\begin{figure}
    \hspace{.5cm}
    \includegraphics[width=0.4\linewidth]{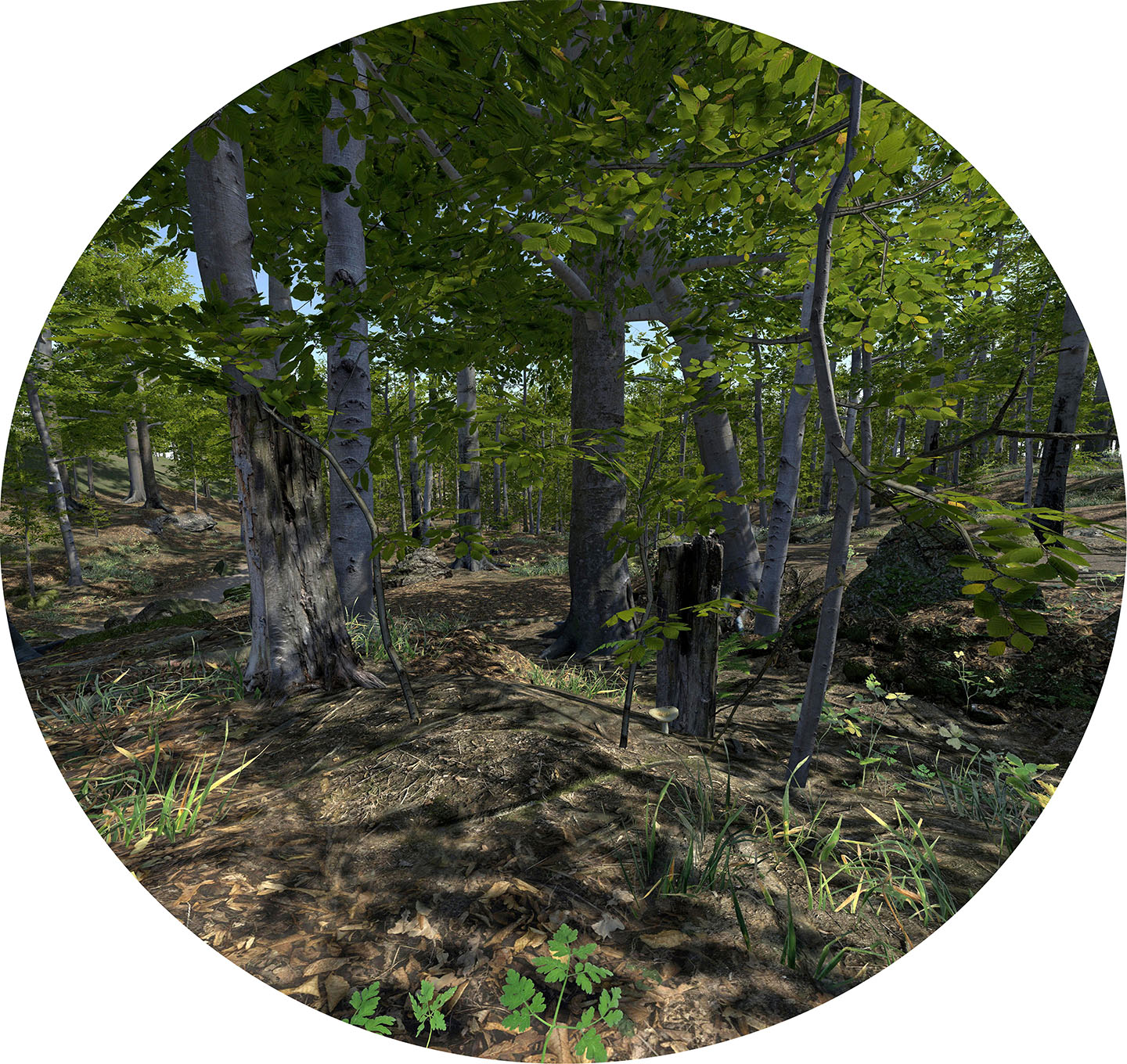}\hfill
    \includegraphics[width=0.4\linewidth]{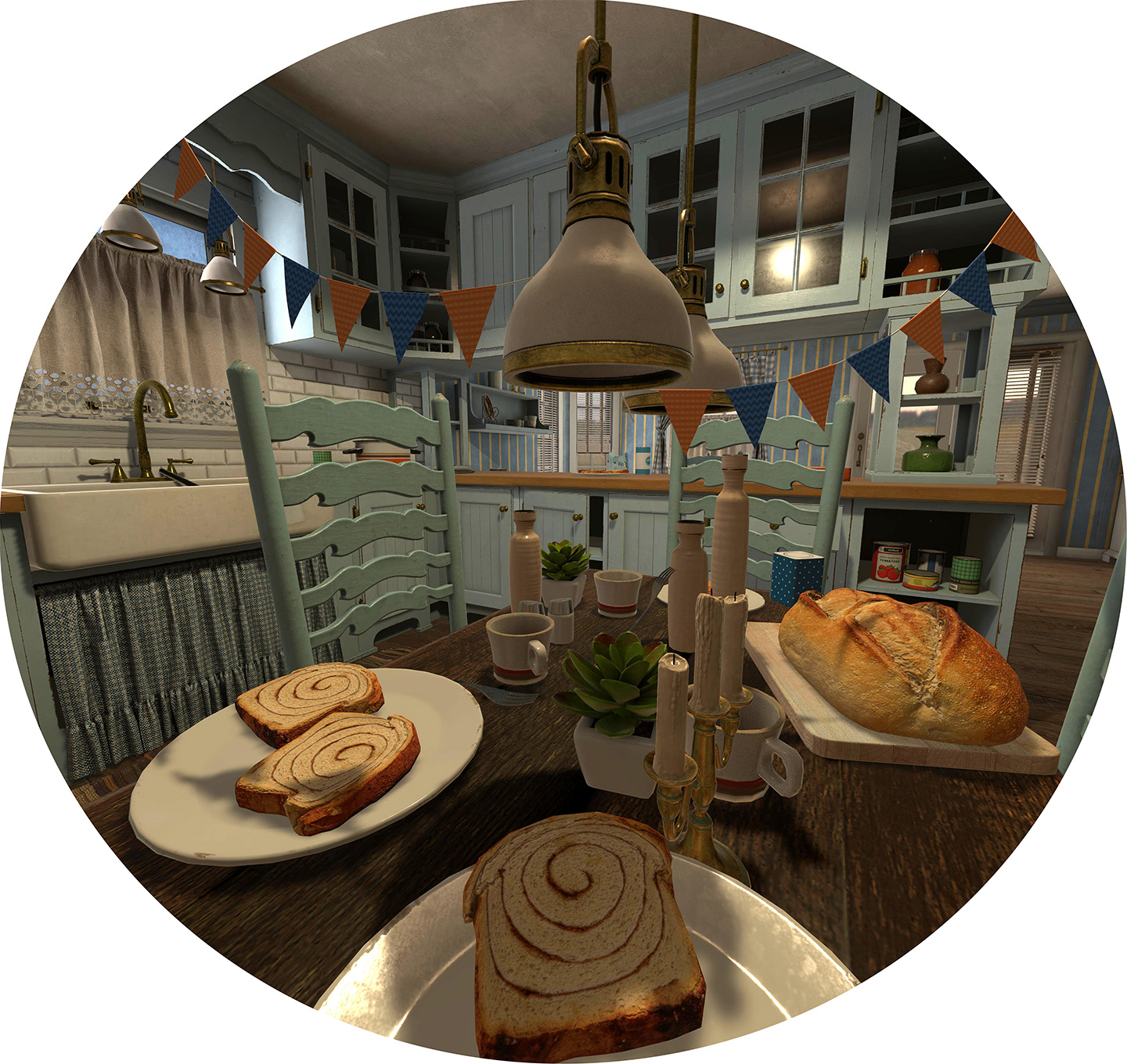}
    \hspace{.5cm}
    
    \caption[Renderings of the two natural scenes (Forest and Kitchen) used in the validation experiment.]
    {Renderings of the two natural scenes (\textsc{Forest}\footnotemark[2] and \textsc{Kitchen}\footnotemark[3]) used in the validation experiment.}
    \label{fig:app-scenes}
    \Description{}
\end{figure}

The experiment described in the previous section was conducted on simple, isolated stimuli. 
To verify whether the model holds for complex stimuli, we conducted a validation experiment on natural scenes and verify that the foveation derived from our model does not affect depth perception. The experiment is set in a \textsc{Forest} and a \textsc{Kitchen} (Figure \ref{fig:app-scenes}).

\footnotetext[2]{\href{www.assetstore.unity.com/packages/3d/vegetation/forest-environment-dynamic-nature-150668}{Forest Environment 
 from NatureManufacture (Unity)}}
\footnotetext[3]{\href{www.assetstore.unity.com/packages/3d/environments/urban/hq-retro-farmhouse-modular-154929}{Retro Farmhouse from NOT\_Lonely (Unity)}}

\subsection{Methods}\label{sec:app-exp1}

We show participants half-split images of the natural scenes.
The central part is blacked out up to $15^\circ$ eccentricity.
We validate our model on high eccentricities, as the fovea remains almost unaffected by our model, yielding no measurable changes.
As we allow the gaze to move within a margin of 3--5$^\circ$ from the fixation point, the effective blacking is \textasciitilde10$^\circ$. Furthermore, an examination of $>15^\circ$ eccentricity retains enough binocular image space, as the headset only supports binocular vision up to $25^\circ$ eccentricity. The left and right halves of the image display different disparity levels: one half is rendered with 0mm \gls{ipd} (no disparity), while the other half is rendered with a varying \gls{ipd} (0--20mm) to control the overall magnitude of disparities in the scene (see Supplementary Material for vergence angle histograms). The \gls{ipd}, and thus the overall scene depth, was adjusted in a staircase procedure. We simulate foveation by locally interpolating between pre-blurred images according to the per-pixel eccentricity value. Afterwards, we linearly map the values of the foveated image to match the value range of the input image.

Participants were shown both scenes in full-resolution (\textsc{Org}) and foveated (\textsc{Fov}) rendering styles. Each participant first viewed the \textsc{Forest} scene (\textsc{F}), and then the \textsc{Kitchen} scene (\textsc{K}). The order of presented rendering styles was randomized, and the style sequence was counterbalanced across scenes (e.g. \textsc{F-Fov}, \textsc{F-Org}, \textsc{K-Org}, \textsc{K-Fov}). Before threshold estimation began, participants were allowed to freely explore each scene using their central vision to familiarize themselves with prominent depth cues. This step was added to reduce measurement noise caused by initial variability in scene familiarity, as identified in pilot testing. Participants were asked to identify the side of the image which exhibited disparity. To estimate the \gls{ipd} threshold, the displayed disparity was adjusted using the "Best PEST" procedure, as described in Section \ref{sec:met-procedure}. The experiment was conducted using the same apparatus as described in Section \ref{sec:met-apparatus}. We conducted the study on 15 participants (13M, 1F, 1X; mean age 26.1 ± 2.5 years). 11 participants reported prior experience with perceptual studies. All had normal or corrected-to-normal vision, signed their previous consent, and were compensated for their time. The study was approved by the institutional ethics committee.

\subsection{Results}

\begin{figure}
    \includegraphics[width=\linewidth]{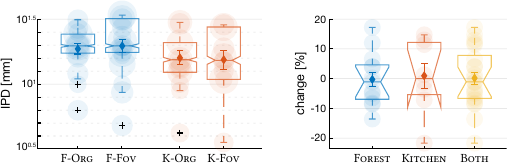}
    \caption{Left: Estimated \gls{ipd} thresholds for both scenes and rendering styles. The area of each dot denotes its weight. The scatter marks the mean and standard error per condition, while the box plots show the median, 25\% and 75\% percentiles, 1.5 IQR whiskers, and 95\% confidence interval. Right: Change of the log-threshold per participant and scene. Only participants with two valid measurements per scene were considered. A positive value denotes a smaller threshold for the \textsc{Fov} style.}
    \label{fig:app-disparity-results}
    \Description{}
\end{figure}

The PEST procedures are evaluated and weighed as stated in Section \ref{sec:mod-threshold-estimation}. Figure \ref{fig:app-disparity-results} shows the estimated \gls{ipd} thresholds, as well as a representation of the change of the participant's \gls{ipd} threshold between the \textsc{Org} and \textsc{Fov} stimuli. The change rate is determined by dividing a participant's threshold presented for an \textsc{Org} stimulus by the threshold of the respective \textsc{Fov} stimulus. This results in a positive change value in case the participant had a lower threshold in the \textsc{Fov} condition.

Neither the means nor the medians differ substantially for both the thresholds and change rates. The overlapping box plot notches yield a 95\% certainty that the medians do not differ significantly, which is the case for all reported values. This suggests that participants exhibited similar stereoacuity for both rendering styles, and that the foveation did not compromise it. This finding is reflected by the constant change rate of \textasciitilde0\%, which shows that the participants' performance did not change in either of the settings. Although the change rates vary up to $\pm 20\%$, we did not find a significant trend in personal preferences, meaning that participants did not have a similar change rate across conditions. In addition, the directions of change were not related to expert vs. inexperienced viewer.

\subsection{Discussion}
The results demonstrate that the removal of high frequencies does not affect stereoacuity. This corresponds to our hypothesis, that high frequencies do not carry essential information for stereo matching. Additionally, the results confirm that our model holds for complex scenes that differ significantly from the simple stimulus used in our first experiment. However, for complex stimuli, we were not able to replicate the increased stereoacuity anticipated with our suggested blur intensity. Two factors may contribute to this: Firstly, the observed decline in $T$ towards $p_2$ was subtle. Secondly, the measurement process for the \gls{ipd} thresholds is inherently more variable due to the abstract nature of the task and the complex stimuli, which may have posed a challenge for participants to resolve consistently. However, the applied foveation did not detrimentally affect overall stereoacuity. 
\section{Future Work}

Recreating a correct spatial understanding is crucial for various real-time applications, such as simulations, medical training, entertainment, and design interfaces. Our findings indicate that strong peripheral blur does not compromise depth perception derived from stereopsis. Furthermore, the results of our psychophysical experiment suggest that an optimal amount of peripheral blur can actually improve stereoacuity. We hypothesize that this improvement may be due to the reduction of spatial aliasing in peripheral vision. However, this argument does not fully account for our observations, as the limits presented by \citeauthor{thibos-1987a} [\citeyear{thibos-1987a}] do not align with the blur levels at which stereoacuity was highest in our experiments.

We believe that further investigation of this effect is a fascinating avenue for future research, which could enhance our understanding of the mechanisms behind depth perception.
This may lead to interesting applications where careful manipulation of image content improves spatial awareness. Additionally, potential findings can lead to new conclusions about the quality and reproduction of visual cues, particularly in light of recent efforts to enhance and recreate high-frequency details in foveated rendering \cite{patney-2016a,Walton2021BeyondB, tariq-2022a}.

It has been demonstrated that stereoacuity decreases in peripheral vision \cite{prince-1998, mochizuki-2012a,blakemore-1970a}. Our results align with these findings and complement them by showing that this trend is not further enhanced by peripheral blur. This highlights the need for careful treatment of depth information in foveated content: Our findings indicate that reducing the spatial quality of an image does not entail that the quality of depth reproduction can be reduced as well. This observation is particularly important in applications that require depth processing. For example, stereoscopic and multiscopic content can be transmitted as a combination of rendered images and depth maps to recreate missing views during the decoding step \cite{merkle2009}. When developing gaze-contingent compression methods, the compression of depth maps should be handled carefully and ideally separately from the compression of the image content. Similarly, any depth-related rendering process, such as displacement mapping or level-of-detail methods \cite{murphy2001,surace-2023a}, should be handled carefully and without implicitly assuming that foveation masks na\"{\i}ve compression of depth information. Exploring these issues in the future will broaden our understanding of the critical factors that must be replicated to achieve real-time realism in virtual and augmented realities.
\section{Conclusion}

Techniques such as foveated rendering are an essential tool for reproducing real-time visual realism in its entirety. However, it is paramount that other dimensions of realism, such as the perception of depth, are maintained in conjunction with perceived spatial realism. The goal of this work is to understand how foveated rendering affects our perception of disparity in immersive environments. To this end, we measure disparity thresholds across various eccentricities (0--20$^\circ$) under varying blur intensities (0--$26.6'$ Gaussian filter $\sigma$). Based on our measurements, we derive a perceptual model which demonstrates that the blur intensities applied in common foveation procedures do not affect stereoacuity. Moreover, the threshold for stable stereoacuity is much higher (\textasciitilde2$\times$) than the blur applied in traditional foveated rendering. We validate these findings on complex scenes and show that the overall perceived depth in a scene is not affected by strong peripheral blur. Additionally, our findings suggest that it is important to maintain high depth quality even in strongly foveated stereoscopic content.

\begin{acks}
This project has received funding from the European Research Council under the European Union’s Horizon 2020 research and innovation program (Grant 804226, PERDY).
\end{acks}

\bibliographystyle{ACM-Reference-Format}
\bibliography{literature}

\end{document}